\documentclass[10pt, a4paper]{article}

\usepackage{lrec2026}  
\usepackage{booktabs}
\usepackage{multirow}
\usepackage{amsmath}
\usepackage{url}
\usepackage{fancyhdr}
\title{How Much Does Persuasion Strategy Matter?\\LLM-Annotated Evidence from Charitable Donation Dialogues}

\name{Tatiana Petrova\thanks{To appear in \textit{Proceedings of the Workshop on Social Context and Integrating NLP and Psychology to Study Social Interactions (SoCon-NLPSI)}, co-located with the 15th Language Resources and Evaluation Conference (LREC 2026), Palma de Mallorca, Spain, May 2026.}, Stanislav Sokol, Radu State}
\address{Interdisciplinary Centre for Security, Reliability and Trust (SnT) \\
         University of Luxembourg \\
         \texttt{\{tatiana.petrova, radu.state\}@uni.lu, stanislav.sokol.001@student.uni.lu}}

\abstract{
Which persuasion strategies, if any, are associated with donation compliance? Answering this requires fine-grained strategy labels across a full corpus and statistical tests corrected for multiple comparisons. We annotate all 10,600 persuader turns in the 1,017-dialogue PersuasionForGood corpus \citep{wang2019persuasion}, where donation outcomes are directly observable, with a taxonomy of 41 strategies in 11 categories, using three open-source large language models (LLMs; Qwen3:30b, Mistral-Small-3.2, Phi-4). Strategy categories alone explain little variance in donation outcome (pseudo $R^2 \approx 0.015$, consistent across all three annotators). Guilt Induction is the only strategy significantly associated with \emph{lower} donation rates ($\Delta \approx -23$ percentage points), an effect that replicates across all three models despite only moderate inter-model agreement. Reciprocity is the most robust positive correlate. Target sentiment and interest predict whether a donation occurs but show at most a weak correlation with donation amount. These findings suggest that strategy identification alone is insufficient to explain persuasion effectiveness, and that guilt-based appeals may be counterproductive in prosocial settings. We release the fully annotated corpus as a public resource.
\\ \newline }

\begin{document}

\maketitleabstract

\section{Introduction}

Charitable donation conversations are a natural setting for studying persuasion: a persuader attempts to convince a target to donate money, and the outcome (donated or not, and how much) is directly observable. The PersuasionForGood corpus \citep{wang2019persuasion}\footnote{We use the publicly available PersuasionForGood dataset \citeplanguageresource{PersuasionForGood}.} provides 1,017 such dialogues collected via Amazon Mechanical Turk, where persuaders try to convince targets to donate part of their task earnings to Save the Children. Understanding which strategies help or hinder can inform the design of prosocial dialogue systems and evidence-based training for charitable fundraisers.

Subsequent work has improved strategy classification accuracy \citep{saha2021persuasion} and analyzed target resistance \citep{tian2020understanding}, but the question of which strategies, if any, are statistically associated with donation outcomes remains only partially addressed. \citet{wang2019persuasion} tested strategy--donation associations via logistic regression on 252 annotated dialogues with 10 strategies, finding only ``Donation information'' significant ($p < 0.05$) without multiple-comparison correction. The limited sample (29\% of the corpus), 10-strategy scheme, and absence of correction for multiple testing leave room for a more comprehensive analysis. To our knowledge, no prior work has tested individual strategy--outcome associations across the full corpus with a fine-grained taxonomy and multiple-comparison corrections.

We address this gap by (1) defining a hierarchical taxonomy of 41 persuasion strategies in 11 categories, grounded in Cialdini's principles of influence \citeyearpar{cialdini1984influence} and Marwell and Schmitt's compliance-gaining strategies \citeyearpar{marwell1967dimensions}; (2) annotating all 10,600 persuader turns using three open-source LLMs (Qwen3:30b as primary annotator, Mistral-Small-3.2 and Phi-4 as robustness checks) and all 10,332 target turns using Qwen3:30b alone (for sentiment and interest labels), following current best practice \citep{carlson2025llm,abdurahman2025primer}; and (3) conducting bivariate tests with multiple-comparison corrections and multivariate logistic regression to assess strategy--donation associations. Our contributions include corpus-scale evidence on strategy--donation associations (including the Guilt Induction backfire and the limited predictive power of strategy categories), a three-model robustness design, and a fully annotated resource covering all 1,017 dialogues.

\section{Related Work}

\paragraph{PersuasionForGood and persuasion in NLP.} \citet{wang2019persuasion} introduced the corpus with 10 strategy labels (e.g., logical appeal, emotional appeal, credibility appeal) and an RCNN-based classifier. \citet{saha2021persuasion} improved classification with BERT-based models, while \citet{tian2020understanding} analyzed target resistance strategies. \citet{chen2021persuasion} proposed weakly supervised identification of 8 persuasion strategies in online contexts.

\paragraph{Persuasion strategy taxonomies.} Our taxonomy builds on Cialdini's \citeyearpar{cialdini1984influence} six principles of influence and Marwell and Schmitt's \citeyearpar{marwell1967dimensions} 16 compliance-gaining strategies, organizing 41 strategies into 11 categories to enable analysis at both category and individual strategy level (Section~3.1).

\paragraph{LLM-as-annotator.} Recent guidance recommends testing multiple models and treating model choice as a researcher degree of freedom \citep{gilardi2023chatgpt,pangakis2024automated,carlson2025llm,abdurahman2025primer}; we follow this practice (Section~3.2).

\section{Methodology}

\subsection{Taxonomy}

Starting from Cialdini's \citeyearpar{cialdini1984influence} principles of influence and Marwell and Schmitt's \citeyearpar{marwell1967dimensions} compliance-gaining strategies, supplemented by work on fear appeals, framing, and emotional manipulation, we compiled 45 candidate strategies. Pilot annotation revealed that several were poorly distinguishable (e.g., overlapping moral and value-based appeals). After iterative merging and refinement, we arrived at 41 persuasion strategies in 11 categories, plus 9 conversation management labels (e.g., Greeting, Acknowledgement). Each strategy has a textual definition, characteristic markers, and decision rules (see released code).

The most frequent persuasion category is Norms / Morality / Values ($n = 1{,}331$, 12.6\% of all persuader turns), followed by Rational / Impact Appeal (9.2\%) and Framing \& Presentation (7.3\%). Overtly coercive strategies (Threat / Pressure, Urgency / Scarcity) are nearly absent, together accounting for only 0.2\% of turns. Guilt Induction, while psychologically manipulative, is placed under Norms / Morality / Values because it operates through moral obligation rather than direct coercion.

\begin{figure}[!ht]
\centering
\includegraphics[width=\columnwidth]{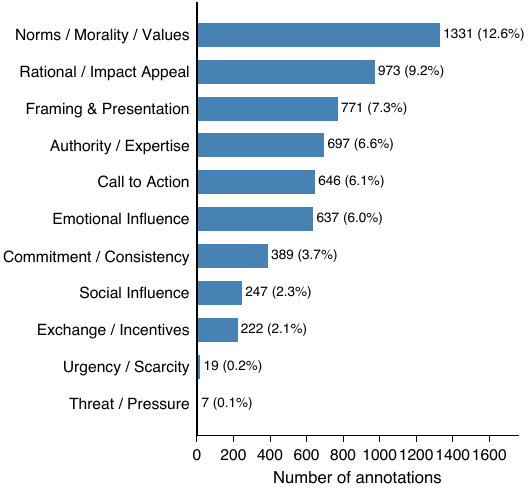}
\caption{Distribution of 11 persuasion strategy categories (Qwen3:30b). Each bar shows the number of persuader turns assigned to the category; percentages indicate the category's share of all $N{=}10{,}600$ persuader turns. Conversation Management turns (44.0\%) are omitted.}
\label{fig:distribution}
\end{figure}

\subsection{Annotation Procedure}

We annotate all 10,600 persuader utterances across 1,017 dialogues using three open-source LLMs deployed locally via Ollama: \textbf{Qwen3:30b} (Alibaba; primary annotator), \textbf{Mistral-Small-3.2} (Mistral AI), and \textbf{Phi-4} (Microsoft, 14B).
%
These models were selected on three criteria.
First, they represent distinct developer families (Alibaba, Mistral AI, Microsoft), reducing the risk that findings reflect idiosyncratic biases of a single training pipeline, in line with the robustness-check methodology of \citet{carlson2025llm} and \citet{abdurahman2025primer}: one model serves as the primary annotator and the others as independent replications.
Second, all three are fully open-weight models deployable locally, ensuring that the annotation pipeline is fully reproducible without dependence on proprietary APIs.
Third, by including models of different parameter scales---Phi-4 (14B) and Qwen3 (30B) as the smallest and largest -- we can assess whether the larger primary annotator's capacity drives results; Table~\ref{tab:replication} shows that key effects hold across model sizes.
Qwen3:30b is designated as primary because it assigns valid labels to all 10,600 turns without errors, produces non-degenerate distributions across all 11 categories (unlike Mistral, which assigns ${<}1\%$ to Framing \& Presentation), and achieves the highest macro-level agreement with \citeauthor{wang2019persuasion}'s gold standard.

All three models use the same two-step hierarchical prompt: the system message instructs the model to act as a ``hierarchical persuasion strategy classification system''; the user message presents (1) the persuader utterance to classify, (2) up to 5 previous dialogue turns as context, and (3) the full category$\to$strategy hierarchy with definitions. The model first selects the parent category, then selects the specific strategy within that category, returning a structured JSON response. Temperature is set to 0.1 for near-deterministic output.

We also annotate all 10,332 target (persuadee) utterances using Qwen3:30b for two dimensions: \emph{sentiment} (negative / neutral / positive, coded as $-1$, $0$, $+1$) and
%
\emph{interest in donation} (not interested / neutral / interested, coded as $0$, $1$, $2$), capturing the target's expressed engagement with the donation topic independently of affective tone: a turn may be affectively neutral yet indicate genuine curiosity about the charity or a willingness to consider donating.
%
We treat mean target sentiment and interest as \emph{covariates} rather than primary predictors: both are measured \emph{during} the conversation and thus reflect the target's evolving response to persuasion rather than pre-existing dispositions; causal direction between strategies, target responses, and donation cannot be established from observational data alone (see Limitations).

\subsection{Annotation Quality}

We evaluate annotation quality at two levels, following best practices for LLM-based annotation \citep{pangakis2024automated}. No human inter-annotator agreement study exists for a 41-label persuasion strategy task; the expected human ceiling for this taxonomy is unknown.

\paragraph{Cross-taxonomy validation.} We compare our annotations against \citeauthor{wang2019persuasion}'s \citeyearpar{wang2019persuasion} gold standard on 300 dialogues (3,047 turns). Because the 41-strategy and 10-strategy schemes are structurally different, we map both to three macro-categories (\emph{persuasive appeal}, \emph{persuasive inquiry}, \emph{non-strategy}), obtaining moderate agreement (Cohen's $\kappa = 0.507$, macro $F_1 = 0.703$; ``moderate'' on the \citealt{landis1977measurement} scale). This is lower than \citeauthor{wang2019persuasion}'s human $\alpha > 0.70$ but was obtained zero-shot without task-specific training.

\paragraph{Expert verification.} An expert in persuasion dialogue reviewed a stratified sample of 100 persuader turns covering all strategy labels annotated by the LLM, confirming correct classification in 84 of 100 cases (84\%). The 16 disagreements predominantly involved boundary cases between semantically adjacent strategies (e.g., Emotional Appeal vs.\ Empathy Appeal, Moral Appeal vs.\ Self-feeling Appeal). As this is verification rather than independent blind annotation, we report accuracy rather than Cohen's $\kappa$.

\paragraph{Inter-model agreement.} On all 10,600 persuader turns, pairwise Cohen's $\kappa$ between the three models ranges from 0.38 to 0.54 at the strategy level (``fair'' to ``moderate'') and from 0.44 to 0.62 at the category level. At the macro level (persuasion vs.\ conversation management), agreement is higher: $\kappa = 0.66$--$0.75$, with raw agreement of 84--88\%. Three-way exact match is 34.1\% for strategies and 47.5\% for categories. The models diverge most on fine-grained labels (e.g., Rational Appeal vs.\ Credibility Appeal) but converge on functional classification. The key downstream findings, in particular the Guilt Induction backfire effect, replicate across all three annotators (Section~\ref{sec:robustness}).

\section{Analysis and Results}
\label{sec:results}

We conduct analyses at two levels of granularity.
At the \textbf{category level} (Section~4.1), we test whether the presence of each of the 11 strategy categories in a dialogue is associated with donation outcome.
At the \textbf{individual strategy level} (Sections~4.2--4.3), we restrict tests to strategies appearing in at least 20 dialogues ($n \geq 20$), a minimum-frequency threshold adopted to ensure reliable chi-square inference; 22 of 41 strategies meet this criterion (full counts in Appendix~\ref{app:taxonomy}).
In both analyses we apply chi-square tests with Bonferroni and Benjamini-Hochberg (FDR) corrections for multiple comparisons, followed by multivariate logistic regression to assess independent effects (Section~\ref{sec:logistic}).
Target sentiment and interest are included as covariates rather than primary predictors, as their role is detailed in Section~3.2; bivariate associations with donation outcome are reported in Section~4.4.

The overall donation rate is 53.6\% (545 / 1,017 dialogues), with a mean donation of \$2.17 (\$4.05 among donors only). Dialogues contain an average of ${\sim}10$ persuader turns, typically employing 4--5 distinct persuasion strategies per dialogue (mean $= 4.4$, $SD = 1.6$, range 0--10); persuasion in this setting is thus multi-strategy by nature. All results below use Qwen3:30b (primary annotator) unless otherwise noted.

\subsection{Strategy Categories Do Not Independently Predict Donation}
\label{sec:categories}

We test whether the presence of each strategy category in a dialogue is associated with donation outcome using chi-square tests with Bonferroni correction for 11 comparisons. None of the 11 strategy categories reach statistical significance (all $p_{Bonf} > 0.05$). The closest is Commitment / Consistency ($p_{Bonf} = 0.146$), followed by Rational / Impact Appeal ($p_{Bonf} = 0.280$). This null result replicates across all three annotators (Table~\ref{tab:replication}).

Because dialogues are long (avg.\ ${\sim}10$ turns) and contain many strategies simultaneously, most categories appear in most dialogues (Norms / Morality / Values in 72\%, Rational / Impact Appeal in 56\%), making binary presence/absence a coarse signal.

\subsection{Guilt Induction is Associated with Lower Donation}
\label{sec:guilt}

At the individual strategy level, we test 22 strategies with $n \geq 20$ dialogues, applying both Bonferroni and Benjamini-Hochberg (FDR) corrections. Guilt Induction is the only strategy significantly associated with \emph{lower} donation likelihood (Table~\ref{tab:results}). Dialogues containing Guilt Induction ($n = 104$) have a 32.7\% donation rate, compared to 56.0\% for dialogues without it ($n = 913$; $\chi^2 = 19.4$, $\phi = 0.14$, $p_{Bonf} < 0.001$, $p_{FDR} < 0.001$). The mean donation amount is also 3.5$\times$ lower (\$0.67 vs.\ \$2.34, Mann-Whitney U $p < 0.001$). This effect replicates across all three annotators: $\Delta = -23.0$~pp with Mistral-Small-3.2 and $-23.9$~pp with Phi-4, both significant at $p_{raw} < 0.05$ (Table~\ref{tab:replication}).

To illustrate, a typical Guilt Induction utterance from a non-donated dialogue reads: \emph{``Kids are dying from hunger every minute. Don't you want to help stop that?''} In contrast, a Reciprocity utterance from a donated dialogue: \emph{``That's great! Every bit helps, I will match your donation myself.''} Guilt induction threatens the target's autonomy; reciprocity creates a mutual exchange.

\begin{figure*}[!ht]
\centering
\includegraphics[width=\textwidth]{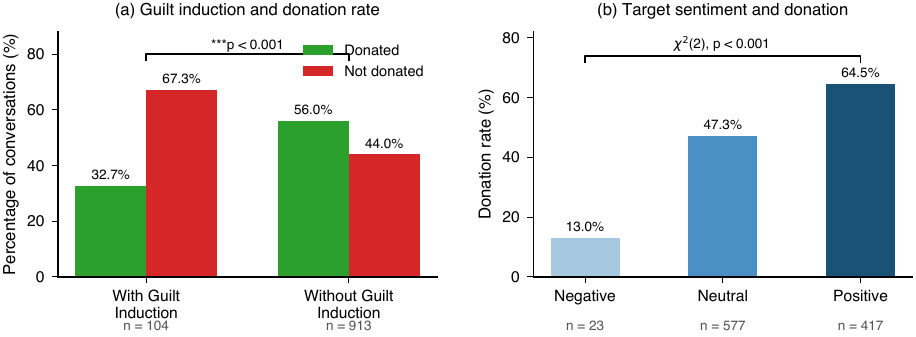}
\caption{(a) Donation rates in dialogues containing Guilt Induction ($n{=}104$) vs.\ dialogues without ($n{=}913$); difference of $-23.3$ pp ($\chi^2{=}19.4$, $p < 0.001$). This effect replicates across all three annotators (Table~\ref{tab:replication}). (b) Donation rate by predominant target sentiment across a dialogue's target turns ($\chi^2(2){=}44.3$, $V{=}0.21$, $p < 0.001$).}
\label{fig:findings}
\end{figure*}

This pattern is consistent with psychological reactance theory \citep{brehm1966theory}: when individuals perceive that their freedom of choice is threatened, they resist rather than comply (Figure~\ref{fig:findings}a).

This association is correlational and may partly reflect reverse causality (see Discussion). Our logistic regression (Section~\ref{sec:logistic}) shows that Guilt Induction remains a negative predictor (odds ratio [OR] $= 0.60$, $p = 0.029$) even when controlling for target sentiment and interest.

\subsection{Reciprocity and Commitment/Consistency are Positive Predictors}
\label{sec:reciprocity}

Reciprocity is the strategy most robustly associated with \emph{higher} donation rates: 72.2\% of dialogues containing Reciprocity ($n = 72$) result in donation, vs.\ 52.2\% without ($n = 945$; $\chi^2 = 10.0$, $\phi = 0.10$, $p_{Bonf} = 0.034$, $p_{FDR} = 0.016$). The positive direction replicates in all three models ($\Delta = +5.8$ to $+20.1$~pp), reaching significance in two of three (Table~\ref{tab:replication}). Commitment and Consistency, while not significant under Bonferroni correction, also reaches significance under FDR ($p_{FDR} = 0.016$; 63.5\% vs.\ 51.2\%). For both strategies, reverse causality cannot be excluded: persuaders may deploy them after the target has already signaled willingness.

\begin{table}[!ht]
\centering
\footnotesize
\setlength{\tabcolsep}{3pt}
\begin{tabular}{@{}lcccc@{}}
\toprule
 & \textbf{Pres.} & \textbf{Abs.} & $\boldsymbol{p_B}$ & $\boldsymbol{p_{FDR}}$ \\
\midrule
\multicolumn{5}{@{}l}{\textit{Strategy--donation (22 tests, $\chi^2$)}} \\
Guilt         & 32.7\% & 56.0\% & ${<}.001$ & ${<}.001$ \\
Reciprocity   & 72.2\% & 52.2\% & .034 & .016 \\
Commit./Cons. & 63.5\% & 51.2\% & .049 & .016 \\
\midrule
\multicolumn{5}{@{}l}{\textit{Target response--donation (MWU)}} \\
 & \textbf{Don.} & \textbf{No don.} & \multicolumn{2}{c}{$\boldsymbol{p}$} \\
Sentiment & $.44$ & $.27$ & \multicolumn{2}{c}{${<}.001$} \\
Interest  & $1.24$ & $1.09$ & \multicolumn{2}{c}{${<}.001$} \\
\midrule
\multicolumn{5}{@{}l}{\textit{Target response--amount (Pearson)}} \\
 & \multicolumn{2}{c}{$\boldsymbol{r}$} & \multicolumn{2}{c}{$\boldsymbol{p}$} \\
Sentiment & \multicolumn{2}{c}{.034} & \multicolumn{2}{c}{.424} \\
Interest  & \multicolumn{2}{c}{${-}.001$} & \multicolumn{2}{c}{.981} \\
\bottomrule
\end{tabular}
\caption{Bivariate associations with donation outcome (Qwen3:30b). \textit{Top}: donation rates (\%) in dialogues where the strategy is present (Pres.) vs.\ absent (Abs.); $p_B$: Bonferroni, $p_{FDR}$: Benjamini-Hochberg correction over 22 tests. Effect sizes are small ($\phi = 0.14$ for Guilt, $0.10$ for Reciprocity). \textit{Middle}: mean target sentiment ($-1$ to $+1$) and interest ($0$--$2$) in donated (Don.) vs.\ not-donated (No don.) dialogues, Mann-Whitney $U$ test. \textit{Bottom}: Pearson $r$ between target response and donation amount (donors only). Cross-model robustness is reported in Table~\ref{tab:replication}.}
\label{tab:results}
\end{table}

\subsection{Target Sentiment and Interest Predict Donation but Not Amount}
\label{sec:sentiment}

The target's expressed sentiment and interest are the variables most strongly associated with whether a donation occurs (both $p < 0.001$; Table~\ref{tab:results}, Figure~\ref{fig:findings}b). Donated dialogues have higher mean target sentiment (0.44 vs.\ 0.27 on a $-1$ to $+1$ scale) and higher mean interest (1.24 vs.\ 1.09 on a 0 to 2 scale). At the turn level, donated dialogues contain more positive target turns (49.5\% vs.\ 38.6\%) and fewer negative turns (6.1\% vs.\ 12.0\%).

However, neither sentiment nor interest shows a statistically significant linear correlation with the donation \emph{amount} among those who did donate (Pearson $r = 0.034$, $p = 0.424$ and $r = -0.001$, $p = 0.981$, respectively). Spearman rank correlation detects a small monotonic association for sentiment ($\rho = 0.112$, $p = 0.009$) but not for interest ($\rho = 0.059$, $p = 0.172$), suggesting a weak non-linear link between target sentiment and donation amount that the linear measure misses. The ``Negative'' sentiment group contains only 23 dialogues (3 donations), so the 13.0\% rate in Figure~\ref{fig:findings}b carries a wide confidence interval and should be interpreted with caution.

\subsection{Logistic Regression: Multivariate Analysis}
\label{sec:logistic}

The bivariate tests above examine each predictor in isolation. To assess whether strategy effects survive when controlling for other predictors, we fit logistic regression models with donation (binary) as the dependent variable (Table~\ref{tab:logistic}).

\paragraph{Model 1: Strategy categories only.} With the 9 strategy categories present in $\geq 20$ dialogues as binary predictors, the model is significant overall (log-likelihood ratio [LLR] $p = 0.011$) but explains little variance (pseudo $R^2 = 0.015$). Commitment / Consistency (OR $= 1.41$, $p = 0.014$) and Rational / Impact Appeal (OR $= 1.36$, $p = 0.018$) are positive predictors; no category is a significant negative predictor.

\paragraph{Model 2: Categories + sentiment + interest.} Adding mean target sentiment and interest improves fit substantially (pseudo $R^2 = 0.082$; likelihood ratio test vs.\ Model~1: $\chi^2 = 94.4$, $p < 10^{-6}$). Sentiment (OR $= 4.24$) and interest (OR $= 3.11$) yield the largest effects (both $p < 0.001$). Commitment / Consistency loses significance ($p = 0.014 \to 0.454$), suggesting either confounding or mediation through target sentiment. Call to Action emerges as a negative predictor (OR $= 0.70$, $p = 0.009$).

\paragraph{Model 3: Parsimonious model (exploratory).} As an exploratory check, we fit a compact model with only Guilt Induction, Reciprocity, mean sentiment, and mean interest (the variables with the strongest bivariate signals). This model achieves nearly the same fit (pseudo $R^2 = 0.080$, Akaike information criterion [AIC] $= 1302$, area under the ROC curve [AUC] $= 0.67$) as the full model (AIC $= 1314$). All four predictors are significant: sentiment (OR $= 3.94$, $p < 0.001$), interest (OR $= 2.77$, $p < 0.001$), Reciprocity (OR $= 2.41$, $p = 0.002$), and Guilt Induction (OR $= 0.60$, $p = 0.029$). For comparison, a categories-only model achieves AUC $= 0.58$. 
\begin{table}[!ht]
\centering
\small
\begin{tabular}{@{}lccc@{}}
\toprule
\textbf{Predictor} & \textbf{OR} & \textbf{95\% CI} & $\boldsymbol{p}$ \\
\midrule
\multicolumn{4}{@{}l}{\textit{Model 3 (parsimonious): pseudo $R^2 = 0.080$, AUC $= 0.67$}} \\
Sentiment     & 3.94 & [2.30, 6.74] & ${<}.001$ \\
Interest      & 2.77 & [1.64, 4.67] & ${<}.001$ \\
Reciprocity   & 2.41 & [1.38, 4.20] & .002 \\
Guilt Ind.    & 0.60 & [0.37, 0.95] & .029 \\
\bottomrule
\end{tabular}
\caption{Logistic regression predicting donation (binary), parsimonious Model~3. OR: odds ratio ($>1$ = higher donation probability). Sentiment: mean target sentiment per dialogue ($-1$ to $+1$). Interest: mean target interest per dialogue ($0$--$2$).}
\label{tab:logistic}
\end{table}

\subsection{Strategy--Response Sentiment Link}

To explore why certain strategies are associated with donation, we pair each persuader turn carrying a persuasion strategy label with the immediately following target turn ($n = 5{,}387$ persuader--target pairs) and compute the mean target sentiment elicited by each strategy (Table~\ref{tab:sentiment-link}). The corpus-wide average response sentiment is $+0.37$. Guilt Induction ($+0.02$) elicits responses far below the corpus average, as do Fear Appeal ($+0.19$) and Unity ($+0.23$). In contrast, Commitment and Consistency ($+0.54$), Reciprocity ($+0.51$), and Foot-in-the-door ($+0.51$) elicit the most positive responses. Reciprocity co-occurs with positive engagement; the affective response pattern points to a possible mediating role in the strategy--donation link, though formal mediation analysis would be needed to confirm this.

\begin{table}[!ht]
\centering
\footnotesize
\setlength{\tabcolsep}{3pt}
\begin{tabular}{@{}lrcr@{}}
\toprule
\textbf{Strategy} & \textbf{n} & \textbf{Mean sent.} & \textbf{\% neg.} \\
\midrule
\multicolumn{4}{@{}l}{\textit{Bottom 3 (lowest mean sentiment)}} \\
Guilt Induction     & 114 & $+0.02$ & 27.2\% \\
Fear Appeal         &  69 & $+0.19$ & 21.7\% \\
Unity               &  43 & $+0.23$ & 23.3\% \\
\midrule
\multicolumn{4}{@{}l}{\textit{Top 3 (highest mean sentiment)}} \\
Commit.\,\&\,Cons.  & 182 & $+0.54$ & 7.1\% \\
Reciprocity         &  67 & $+0.51$ & 7.5\% \\
Foot-in-the-door    &  41 & $+0.51$ & 7.3\% \\
\midrule
\textit{Corpus avg.} & 5,387 & $+0.37$ & --- \\
\bottomrule
\end{tabular}
\caption{Mean target sentiment ($-1$ to $+1$) in the turn immediately following each persuader strategy ($n = 5{,}387$ persuader--target pairs). Only persuasion strategies with $n \geq 20$ pairs shown; Conversation Management labels excluded. \% neg.: proportion of negative target responses.}
\label{tab:sentiment-link}
\end{table}

\subsection{Cross-Model Robustness}
\label{sec:robustness}

To assess whether our findings depend on the choice of annotator, we replicate the full analysis pipeline with Mistral-Small-3.2 and Phi-4 annotations (Table~\ref{tab:replication}). The Guilt Induction backfire effect is the most robust finding: the effect direction and magnitude ($\Delta \approx -23$~pp) are consistent across all three model families, despite the models identifying different numbers of Guilt turns (Qwen: 104 dialogues, Mistral: 35, Phi-4: 36) and achieving only moderate pairwise agreement ($\kappa = 0.38$--$0.54$). To probe whether this consistency reflects a core set of ``obvious'' guilt turns, we examine the intersection: only 16 dialogues are flagged by all three models (Jaccard $= 0.14$), and 63 are flagged by Qwen alone. The donation rate is low for both subsets (37.5\% for all-three-agree, 30.2\% for Qwen-only; Fisher $p = 0.56$ for the difference), indicating that the effect is not confined to extreme cases; even borderline guilt turns identified by a single model are associated with lower donation rates (any-guilt union: 29.9\% vs.\ no-guilt: 56.7\%, $p < 0.001$). The positive Reciprocity association replicates in direction across all three models and reaches significance in two of three. The null result for categories (pseudo $R^2 \approx 0.015$) and the dominance of sentiment and interest in the full model (pseudo $R^2 \approx 0.08$) are stable across all annotators.

\begin{table}[!ht]
\centering
\footnotesize
\setlength{\tabcolsep}{3pt}
\begin{tabular}{@{}lcccc@{}}
\toprule
\textbf{Finding} & \textbf{Qwen} & \textbf{Mistral} & \textbf{Phi-4} & \textbf{Repl.} \\
\midrule
Guilt $\Delta$ (pp) & $-23.3${\tiny ***} & $-23.0${\tiny *} & $-23.9${\tiny **} & 3/3 \\
Recip.\ $\Delta$ (pp) & $+20.1${\tiny **} & $+5.8$ & $+10.4${\tiny *} & 2/3 \\
Categories $R^2$ & .015 & .011 & .016 & 3/3 \\
\;+ Sent./Int.\ $R^2$ & .082 & .079 & .077 & 3/3 \\
\bottomrule
\end{tabular}
\caption{Cross-model robustness. $\Delta$: difference in donation rate (pp) between dialogues with vs.\ without the strategy. $R^2$: McFadden pseudo $R^2$. Significance: $^{*}p{<}.05$, $^{**}p{<}.01$, $^{***}p{<}.001$ (uncorrected $p_{raw}$; multiple-comparison correction is applied only to the primary Qwen model in Table~\ref{tab:results}). Repl.: models where the finding reaches significance ($p_{raw}$) or matches direction (for $R^2$).}
\label{tab:replication}
\end{table}

\section{Discussion and Conclusion}
\label{sec:discussion}

Our results indicate that persuasion effectiveness cannot be reduced to ``strategy X leads to donation.''

First, strategy categories have limited predictive power (pseudo $R^2 = 0.011$--$0.016$ across all three annotators), challenging the assumption that strategy identification alone captures persuasion effectiveness. With 4--5 strategies per dialogue, binary presence/absence is inherently coarse; future work should model strategy \emph{sequences} and \emph{combinations}.

The moderate inter-model agreement ($\kappa = 0.38$--$0.54$) is expected for a 41-label zero-shot task with many semantically adjacent strategy pairs. Key findings replicate regardless of this disagreement (Table~\ref{tab:replication}), while weaker effects, notably Reciprocity (significant in two of three models), should be interpreted with more caution. Following \citet{carlson2025llm}, effects replicating across all three annotators provide stronger evidence than single-model results.

Second, the Guilt Induction backfire effect has a practical implication: prosocial dialogue systems may benefit from avoiding guilt-based appeals. Temporal analysis supports this: splitting the 104 guilt-containing dialogues by the position of the first guilt turn, donation rates decline monotonically from early ($45.9\%$, $n{=}37$) through mid ($34.3\%$, $n{=}35$) to late guilt ($15.6\%$, $n{=}32$), compared to the no-guilt baseline of $56.0\%$ ($\chi^2(3) = 26.7$, $p < 0.001$). After Bonferroni correction, only late guilt differs significantly from the baseline ($p_{adj} < 0.001$). This pattern may reflect reverse causality (persuaders may resort to guilt after sensing resistance), but even early guilt underperforms the baseline numerically.

\paragraph{Resource contribution.} We release the full annotated dataset (20,932 turns across 1,017 dialogues): strategy labels from all three annotators for all 10,600 persuader turns and sentiment/interest labels for all 10,332 target turns, together with prompt templates, analysis scripts, and validation code.\footnote{Code and data: \url{https://github.com/persuasion-nlp/persuasion-strategies} } The full taxonomy is in Appendix~\ref{app:taxonomy}.

\section{Limitations}

Our LLM annotations are produced without fine-tuning; inter-model agreement on fine-grained labels is moderate ($\kappa = 0.38$--$0.54$), and annotation noise may attenuate downstream estimates. Key findings (Guilt backfire, low category $R^2$) are robust to annotator choice, but weaker effects (e.g., Reciprocity) vary across models. Each turn receives one label, though turns may contain multiple strategies; forced single-label annotation may undercount co-occurring strategies. Sentiment and interest are measured \emph{during} the conversation, so they function as concurrent mediators rather than exogenous predictors; causal mediation analysis would be needed to disentangle strategy effects from target response effects. The released data lacks individual worker identifiers, so non-independence across dialogues cannot be ruled out. Finally, all models explain at most 8\% of variance (pseudo $R^2 = 0.08$); all reported associations are correlational. Expert verification was conducted by a single annotator with expertise in persuasion and dialogue research; while the 84\% accuracy on a stratified 100-turn sample suggests acceptable label quality, a multi-annotator blind evaluation would yield a more reliable estimate and is left for future work.

\section{Ethics Statement}

This work analyzes existing publicly available dialogue data \citep{wang2019persuasion}. No new human subjects data was collected. The persuasion strategies we study are from cooperative charitable donation contexts. Findings about persuasion strategy effectiveness could theoretically inform manipulative applications; however, the primary intended use is understanding human persuasion dynamics and improving prosocial dialogue systems.

\section*{List of Abbreviations}

\begin{tabular}{@{}ll@{}}
AIC  & Akaike Information Criterion \\
AUC  & Area Under the ROC Curve \\
CI   & Confidence Interval \\
FDR  & False Discovery Rate \\
LLM  & Large Language Model \\
LLR  & Log-Likelihood Ratio \\
MWU  & Mann-Whitney $U$ (test) \\
OR   & Odds Ratio \\
pp   & percentage points \\
\end{tabular}

\section{Bibliographical References}\label{sec:reference}

\bibliographystyle{lrec2026-natbib}
\bibliography{references}

\section{Language Resource References}
\label{lr:ref}
\bibliographystylelanguageresource{lrec2026-natbib}
\bibliographylanguageresource{languageresource}

\appendix

\section{Full Strategy Taxonomy}
\label{app:taxonomy}

Table~\ref{tab:full-taxonomy} lists all 41 persuasion strategies and 9 conversation management labels with turn counts.

\begin{table*}[!ht]
\centering
\small
\begin{tabular}{@{}llrr@{}}
\toprule
\textbf{Category} & \textbf{Strategy} & \textbf{n} & \textbf{\%} \\
\midrule
\multirow{4}{*}{Norms / Morality / Values}
  & Appeal to Values & 543 & 5.1 \\
  & Moral Appeal & 521 & 4.9 \\
  & Guilt Induction & 129 & 1.2 \\
  & Self-feeling Appeal & 138 & 1.3 \\
\addlinespace
\multirow{2}{*}{Rational / Impact Appeal}
  & Rational Appeal & 928 & 8.8 \\
  & Logical Appeal & 45 & 0.4 \\
\addlinespace
\multirow{4}{*}{Framing \& Presentation}
  & Framing & 751 & 7.1 \\
  & Loss Aversion Appeal & 18 & 0.2 \\
  & Bait-and-switch & 1 & {$<$0.1} \\
  & Pretexting & 1 & {$<$0.1} \\
\addlinespace
\multirow{3}{*}{Authority / Expertise}
  & Credibility Appeal & 683 & 6.4 \\
  & Expertise & 7 & 0.1 \\
  & Authority & 7 & 0.1 \\
\addlinespace
\multirow{6}{*}{Emotional Influence}
  & Empathy Appeal & 265 & 2.5 \\
  & Storytelling & 180 & 1.7 \\
  & Emotional Appeal & 76 & 0.7 \\
  & Fear Appeal & 73 & 0.7 \\
  & Sympathy Appeal & 40 & 0.4 \\
  & Emotional Manipulation & 3 & {$<$0.1} \\
\addlinespace
\multirow{2}{*}{Call to Action}
  & Call to Action & 635 & 6.0 \\
  & Liking & 11 & 0.1 \\
\addlinespace
\multirow{4}{*}{Commitment / Consistency}
  & Commitment and Consistency & 232 & 2.2 \\
  & Activ.\ of Personal Commitment & 95 & 0.9 \\
  & Foot-in-the-door & 45 & 0.4 \\
  & Door-in-the-face & 17 & 0.2 \\
\addlinespace
\multirow{3}{*}{Social Influence}
  & Social Proof & 191 & 1.8 \\
  & Unity & 53 & 0.5 \\
  & Social Positioning & 3 & {$<$0.1} \\
\addlinespace
\multirow{4}{*}{Exchange / Incentives}
  & Reciprocity & 88 & 0.8 \\
  & Rewarding Activity & 71 & 0.7 \\
  & Pre-giving & 57 & 0.5 \\
  & Debt & 6 & {$<$0.1} \\
\addlinespace
\multirow{2}{*}{Urgency / Scarcity}
  & Urgency & 17 & 0.2 \\
  & Scarcity & 2 & {$<$0.1} \\
\addlinespace
\multirow{2}{*}{Threat / Pressure}
  & Threat & 5 & {$<$0.1} \\
  & Aversive Stimulation & 2 & {$<$0.1} \\
\midrule
\multirow{9}{*}{Conversation Management}
  & Greeting / Rapport & 1,060 & 10.0 \\
  & Acknowledgement & 1,022 & 9.6 \\
  & Charity Awareness Probe & 792 & 7.5 \\
  & Non-persuasive Other & 608 & 5.7 \\
  & Logistics / Coordination & 350 & 3.3 \\
  & Donation Baseline / Habit Probe & 325 & 3.1 \\
  & Conversation Closing & 297 & 2.8 \\
  & Qualification / Segmentation & 138 & 1.3 \\
  & Permission / Time Check & 69 & 0.7 \\
\bottomrule
\end{tabular}
\caption{Complete taxonomy: 41 persuasion strategies in 11 categories plus 9 conversation management labels (Qwen3:30b annotations). Columns show absolute turn counts and each label's percentage of all $N{=}10{,}600$ persuader turns (percentages sum to 100\%). Five of the 41 taxonomy strategies (Punishing Activity, Overloading, Confusion Induction, Promise, and Activation of Impersonal Commitment) received zero assignments and are not shown.}
\label{tab:full-taxonomy}
\end{table*}

\end{document}